\documentclass{article}
\usepackage[latin1]{inputenc} 
\usepackage{graphicx}
\usepackage{color}
\usepackage{psfrag}
\usepackage{amsmath}
\usepackage{amssymb}
\usepackage{amsthm}
\usepackage{listings}
\usepackage{bm}
\usepackage{multirow}

\title{Sparsification and feature selection by compressive linear regression}

\author{
Florin Popescu Ph.D.* and Daniel Renz, \\
*Fraunhofer FIRST, Intelligent Data Analysis - IDA, \\ Kekuléstraße 7, 12489 Berlin, Germany \\
}

\DeclareMathOperator*{\argmin}{arg\,min}

\renewcommand{\vec}[1]{\bm{#1}}

\date{ }

\begin{document}

\maketitle

\begin{abstract}
The Minimum Description Length (MDL) principle states that the optimal model for a given data set is that which compresses it best. Due to practial limitations the model can be restricted to a class such as linear regression models, which we address in this study. As in other formulations such as the LASSO and forward step-wise regression we are interested in sparsifying the feature set while preserving generalization ability. We derive a well-principled set of codes for both parameters and error residuals along with smooth approximations to lengths of these codes as to allow gradient-descent optimization of description length, and go on to show that sparsification and feature selection using our approach is faster than the LASSO on several datasets from the UCI and StatLib repositories, with favorable generalization accuracy, while being fully automatic, requiring neither cross-validation nor tuning of regularization hyper-parameters, allowing even for a nonlinear expansion of the feature set followed by sparsification.
\end{abstract}

\section{Introduction}

\subsection{MDL}

The Minimum Description Length (MDL) principle is one of many methods that have been proposed to tackle the general model selection problem through principled balancing of the cost of storing a given model's prediction errors on a given data set and that of storing the model's parameters. Based on algorithmic information theory, it was first formulated by Rissanen in 1978 \cite{rissanen78}. Other methods include AIC \cite{akaike73}, BIC \cite{schwarz78} and MML \cite{wallace87}, which are based on Bayesian statistics and classical information theory. In the past decades several theoretical approaches to MDL have evolved \cite{li89,rissanen78,vitanyi00}, while practical implementations remain an open research topic \cite{schmidhuber02}, as the potential computational burden MDL can impose is significant.

\textit{Practical MDL} uses description methods that are less expressive than universal languages, so that the length of the shortest description of the data is always computable  \cite{li97,schmidhuber02,solomonoff64a,solomonoff64b}. It restricts the set of allowed codes in a manner which allows us to find the shortest code length of the data, relative to the set of allowed codes in finite time. So far, three practical MDL schemes have been devised \cite{lanterman01}. We used the simplest of them, which is called \textit{two-part MDL}: $\hat{M} = \text{argmin}_{M \in \mathcal{M}} \left( L\left(D|M\right) + L\left(M\right) \right)$, where $D$ is the data, $M$ the model, $\mathcal{M}$ the model class (the set of allowed models) and $L$ the code length function. If we knew the probability distributions $p(M), p(D|M)$, we could use  maximum likelihood estimation (the problem being in essence a Bayesian learning formulation) or appropriate asymptotically compact codes, i.e. Shannon-Fano coding, to generate the code books for $L(D|M)$ and $L(M)$ (as it is done in MML). However, we do not know them, so in order to cover a wide set of possible underlying noise probability densities and size of training data sets we used a novel approach based on codes we developed ourselves.

\subsection{Regularized linear regression vs. compressive linear regression}

In linear regression, we assume we have data organized in an observation matrix $\text{D}\in \mathbb{R}^{J\mathrm{\times N}}$
where $N$ is the number of observations and $J$ is the number of features. This matrix may be expanded in two ways: by adding functions of groups of observations (therefore increasing $N$), as in kernel expansion, or adding functions of  feature groups (feature expansion), thereby increasing $J$. In linear regression it is customary to add at least one constant row to the features, corresponding to the bias. Feature expansion can be more generally written as

\begin{equation}
\text{X}\in \left \lbrace {\left \lbrace {\mathbf{b}_{q},f_{q}:q=1,2,3,...,p} \right \rbrace :\mathbf{b}_{q}\subset \{1,2,3,...,J\},f_{q}:\mathbb{R}^{|\mathbf{b}_{q}|}\rightarrow \mathbb{R}^{1}} \right \rbrace   
\end{equation}

Thus $X$, which we call the \textit{feature product}, is a subset of matrix functions $\text{X}\subset \mathbb{R}^{J\times N}\rightarrow \mathbb{R}^{(J+p)\times N}$ operating on groups of rows, as to allow eventual feature selection. The linear regression formulation can be written as

\begin{equation}
\vec{y} = X(D) \vec{\theta} +\vec{e}, \vec{y} \in \mathbb{R}^{N}, \vec{\theta}\in \mathbb{R}^{K}, 
\end{equation}

where $K = J+p$. Depending on our prior knowledge of the probability density function $p(\vec{e})$  (i.e. of measurement errors) there are various ways in which an approximate solution $\vec{\hat{\theta}}$ may be sought, by minimizing the appropriate loss function (or log-likelihood). One of the most common ``non-uniformative'' assumptions, consistent with a maximal entropy distribution for finite variance i.i.d sampling, is that the errors are Gaussian and the corresponding loss function to be minimized is provided by the 2-norm

\begin{equation}
\label{eq:lasso}
 \vec{\hat{\theta}} = \argmin_{\vec{\theta}} \left( \left \| \vec{y} - X(D) \vec{\theta} \right \|_{2}+\gamma \left \|{\vec{\theta }} \right \|_{r} \right)
\end{equation}

The additional $r$-norm is necessary, even without explicit regularization ($\gamma =1$), to cover the underdetermined case ($K>N$), but without regularization $\ell_{2}$ regression can very often overfit \cite{golub89}. Various regularization norms have been proposed, the most commonly used being ridge regression and LASSO \cite{tibshirani96}. The Tikhonov regularization \cite{tikhonov77} proposes a generalized 2-norm, but commonly the straightforward $\ell_{2}$ norm is used (ridge regression). The LASSO proposes the $\ell_{1}$ norm and  offers the additional advantage of a tendency to sparsify $\vec{\theta}$. Both approaches can be seen, from a Bayesian perspective, to assume a Gaussian error distribution $p(\vec{e})$ and independent zero-mean priors with uniform variances over components of $\vec{\theta}$, of Gaussian form for ridge regression and double-exponential for the LASSO. The Bayesian interpretation is problematic however: each feature may have different measurement units (scale), so a likelihood interpretation based on the sum of absolute values of the coefficients (each having inverse units of the corresponding feature) is rather non-informative. Ridge-regression is likewise not scale invariant. Furthermore, the variance of the parameter ``prior'' depends on the hyper-parameter $\gamma$, which is almost always chosen via posterior analysis of the data. Although LASSO and ridge regression, for fixed $\gamma$, consist of convex optimization problems which can be accurately and efficiently solved, they are not convex in $\gamma$: complete model selection requires computationally expensive bootstrapping and cross-validation, while the actual statistic to be minimized depends on the particular implementation.

Our objective was to provide for a scale invariant objective function which allows for meaningful Bayesian interpretation and does not require hyper-parameter tuning. The MDL formulation of regularized regression requires, first of all, the weak assumption that the target has finite and uniform measurement precision or quantization width $\delta(\vec{y})$ (i.e. $ \vec{y}/\delta(\vec{y}) \in \mathbb{Z}^{N}$). We seek

\begin{equation}
\label{eq:mdl}
\min_{\vec{\theta_{\#}}} \left( L\left( \frac{\vec{y}}{\delta(\vec{y})} -\left \lfloor  \frac{ X(D) \vec{\theta_{\#}}}{\delta(\vec{y}) } \right \rfloor  \right) + L\left (\vec{\theta_{\#}} \right) \right )\text{,      where    }\vec{\theta_{\#}} \in \mathbb{Q}^{K}
\end{equation}

The difference between equations \eqref{eq:lasso} and \eqref{eq:mdl} is that we are optimizing over rational-valued vectors $\vec{\theta_{\#}}$, rather than real-valued ones, as both the parameter prior and the measurement error distributions are now discrete probabilities rather than probability density functions. Clearly, a direct optimization over rational numbers would be very difficult. This is precisely why we develop codes, approximations and bounds to make MDL optimization possible for the model class of linear regression models with nonlinear feature products. We shall call this method \textit{Compressive Linear Regression (CLR)}.

\section{Methods}
The basic approach we will follow is to compute, as precisely as possible, description lengths of regression parameters and residuals such that a smooth objective function which approximates the description length of the data can be built. For this purpose, known problems such as compact codes for integers, rationals and random integer sequences must be revisited.

\subsection{Coding of integers}

Rissanen's universal prefix-free code (that is, no code word is a prefix of any other code word) for integers \cite{rissanen84} provides a Bayesian ``non-uniformative'' prior $\text{2}^{-L(i)}$, a principle which can be extended from the integer set to rational numbers as well, which we will propose in this paper.  A universal code is a prefix-free code which has an expected code length over all integers that is invariant, within a constant, no matter what the prior probability of the integers are (given that the latter monotonically decreases). However, Rissanen's code is not very compact as defined by the Kraft inequality: $\sum_{d \in \mathcal{D}} 2^{-L(d)} \leq 1 $. A \textit{compact code} is a code, for which the Kraft sum is close to one, which implies that only few of all possible code words of a certain length are not in the code. For every set of code word lengths that satisfy the inequality, there exists a unique decodable prefix-free code with the same code word lengths \cite{gruenwald00}. This is why we can restrict our attention to prefix-free codes throughout this paper. Another argument against using Rissanen's code is that the area under the curve of the code length function is not convex everywhere, being concave at $x_1 = 2, x_2=2^2=4, x_3=2^4=16, x_4=2^{16}, ...$, which can cause problems for some optimization algorithms.

Universal codes are the basic building block for the other codes described, therefore we chose to build a smooth, compact and convex universal code $U_n$ for non-negative integers (and consequently for the signed integers as well via the ordering 0,-1,1,-2,2,-3 etc. We call the code for the signed integers $U$). Our code uses two different coding schemes: For small numbers, it uses a coding scheme $F$ similar to Fibonacci coding \cite{apostolico87} and then switches at a certain point to a coding scheme $E$ producing code words with the same lengths as Elias-Delta coding \cite{elias75} for large numbers. In $F$ the first two codewords are $0$ and $100$. The following $k$ codewords are produced by binary-adding the number 1. For $k$ we use consecutive Fibonacci numbers, i.e. 1,1,2,3,5,... The $k$-th codeword is additionally appended by ``0''. In $E$, the codeword for a number $n$ consists of the three binaries $t$, $blen$ and $b^*$. $t$ are $len - 1$ ones, $len$ is the length of the binary representation $b(n)$ of $n$, $blen$ is the binary representation of $b(n)$ with the first bit set to zero and $b^*$ is $b(n)$ without the first bit.

\subsection{Coding of rationals or simplest rational within an interval on the real line}

Any universal integer code can be used to construct a prefix-free code for rational numbers of specified precision. However, there is no obvious ordering scheme for rationals in terms of code lengths. Li and Vit\'{a}nyi offer a method to map rationals into the unit interval $\left[ 0,1 \right] \in \mathbb{R}$ using the idea of cylinder sets \cite{li97}. We implement a variant of this code and name it $\alpha$-code: 
\begin{equation}
\alpha \text{ : }\mathbb{R}^{2}\rightarrow \{0,1\}^* \text{  ,    }\alpha \text{ }^{-1}\text{: }\{0,1\}^* \rightarrow \mathbb{Q}\text{ ,     s.t.   }\\
\theta _{\#}=\alpha ^{-1}\left ({\alpha (\theta ,\delta )} \right )\in (\theta -\delta ,\theta -\delta ) )\label{eq:}
\end{equation}

where $\alpha^{-1}(\alpha(\theta,\delta))$ is the rational number coded by the binary $\alpha(\theta,\delta)$. Using the code $\alpha$, we can recover a rational number close to the \textit{real} number $\theta$ to a precision of at least $\delta$. We define $\alpha = \alpha (\theta,\delta) = U(\theta_{\delta})U(\theta_{\log})$, where $\theta$ is the rational number to be encoded, $\delta$ the precision to which it should be encoded, $\theta_{\delta}=\left\lceil \theta / \delta \right\rceil$ and $\theta_{\log} = \left\lceil \log | \theta | \right\rceil - \left\lceil 0.5 \log n \right\rceil$. The decimal value of the code word $\alpha (\theta,\delta)$ is $\theta_{\#} = \theta_{\delta} \cdot 2^{\left\lceil \log \theta_{\delta} \right\rceil + \theta_{\log}}$. As $\alpha$ is prefix-free, the code for a vector of rationals which are not presumed to be interdependent is simply the concatenation of the codes of its elements: $\alpha(\vec{\theta}, \vec{\delta}) = \sum_i \alpha(\theta_i, \delta_i)$. We shall use $\alpha$ to store the linear regression parameters, as detailed below. Through numerical tests, we were able to find a smooth function $\bar{\alpha}$ that approximates our $\alpha$-code lengths, for which we calculated an approximation mean error of 0.8 bit  by sampling $10^5$ evenly spaced points for $\theta \in [2^{-8}; 2^8]$ and a \textit{relative precision} $\theta/\delta \in [2^{-8}; 2^0]$:

\begin{eqnarray}
\bar{\alpha} &=& c_0 \cdot \log \left( \tau(\theta) \cdot \left( \text{erf} \left( 10 \cdot \left( \frac{\theta - \delta}{\theta}\right)+1\right) \cdot \left| \theta \right| + \delta \right) - \log \delta  \right) + 1 \\
\tau(\theta) &=& \left| c_1 \cdot  \left( \log \left( \log \left( \theta^2 + c_2 \right) \right) \right) - 1 \right| \nonumber
\end{eqnarray}

\begin{figure}[t]
  \begin{minipage}[t]{.49\textwidth}
    \begin{center}  
%
%
\begin{psfrags}%
\psfragscanon%
%
\psfrag{s02}[rt][rt]{\color[rgb]{0,0,0}\setlength{\tabcolsep}{0pt}\begin{tabular}{r}\Large log($\theta$)\end{tabular}}%
\psfrag{s03}[lt][lt]{\color[rgb]{0,0,0}\setlength{\tabcolsep}{0pt}\begin{tabular}{l}\Large log($\theta/\delta$)\end{tabular}}%
\psfrag{s04}[b][b]{\color[rgb]{0,0,0}\setlength{\tabcolsep}{0pt}\begin{tabular}{c}\Large code length\end{tabular}}%
%
\psfrag{x01}[t][t]{-8}%
\psfrag{x02}[t][t]{-6}%
\psfrag{x03}[t][t]{-4}%
\psfrag{x04}[t][t]{-2}%
\psfrag{x05}[t][t]{0}%
\psfrag{x06}[t][t]{2}%
\psfrag{x07}[t][t]{4}%
\psfrag{x08}[t][t]{6}%
\psfrag{x09}[t][t]{8}%
%
\psfrag{v01}[r][r]{-10}%
\psfrag{v02}[r][r]{-5}%
\psfrag{v03}[r][r]{0}%
\psfrag{v04}[r][r]{5}%
%
\psfrag{z01}[r][r]{0}%
\psfrag{z02}[r][r]{5}%
\psfrag{z03}[r][r]{10}%
\psfrag{z04}[r][r]{15}%
\psfrag{z05}[r][r]{20}%
\psfrag{z06}[r][r]{25}%
%
\resizebox{6cm}{!}{\includegraphics{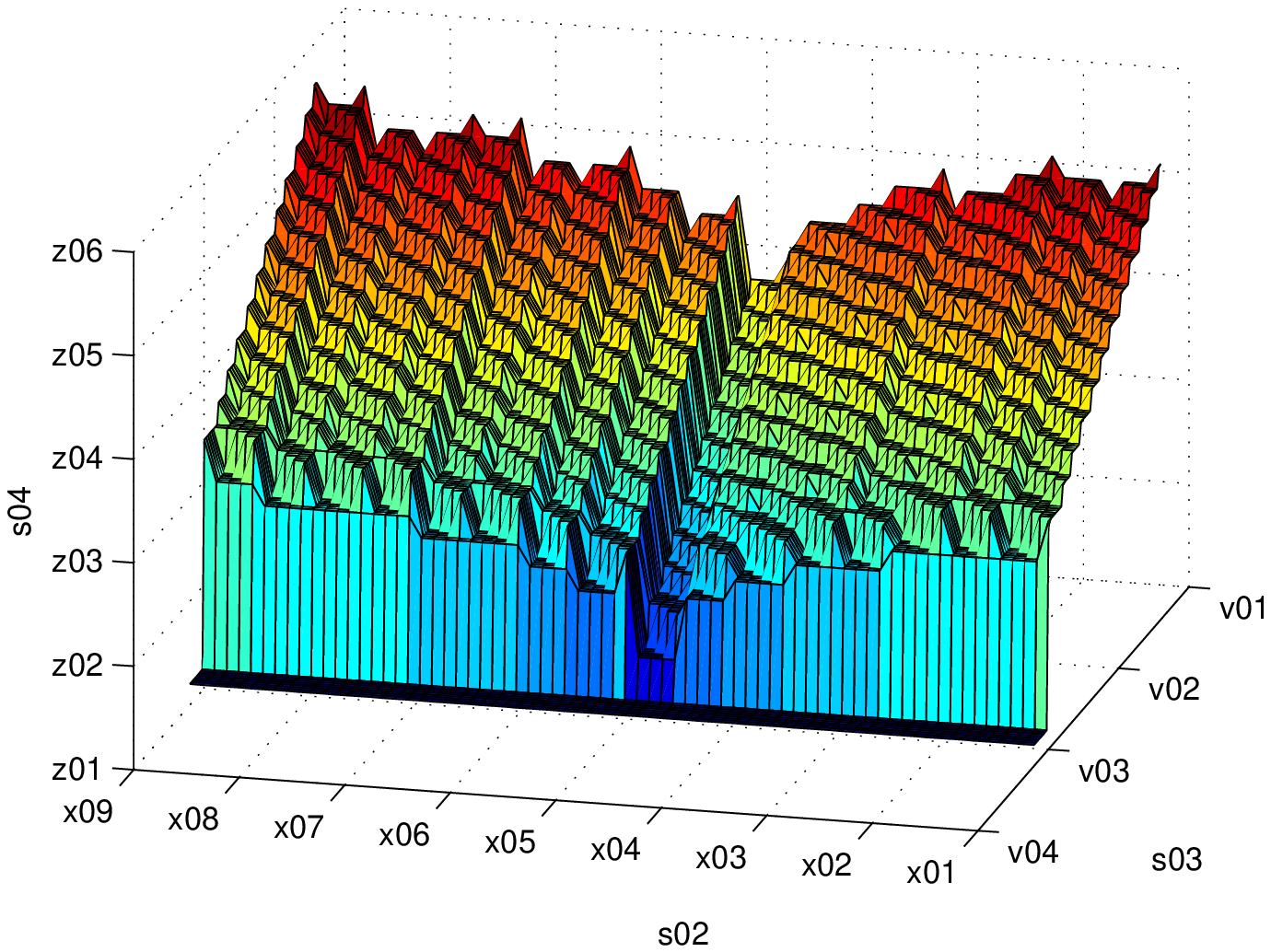}}%
\end{psfrags}%
%

    \end{center}
  \end{minipage}
  \begin{minipage}[t]{.49\textwidth}
    \begin{center}  
%
%
\begin{psfrags}%
\psfragscanon%
%
\psfrag{s02}[rt][rt]{\color[rgb]{0,0,0}\setlength{\tabcolsep}{0pt}\begin{tabular}{r}\Large log($\theta$)\end{tabular}}%
\psfrag{s03}[lt][lt]{\color[rgb]{0,0,0}\setlength{\tabcolsep}{0pt}\begin{tabular}{l}\Large log($\theta/\delta$)\end{tabular}}%
\psfrag{s04}[b][b]{\color[rgb]{0,0,0}\setlength{\tabcolsep}{0pt}\begin{tabular}{c}\Large approx. code length\end{tabular}}%
%
\psfrag{x01}[t][t]{-8}%
\psfrag{x02}[t][t]{-6}%
\psfrag{x03}[t][t]{-4}%
\psfrag{x04}[t][t]{-2}%
\psfrag{x05}[t][t]{0}%
\psfrag{x06}[t][t]{2}%
\psfrag{x07}[t][t]{4}%
\psfrag{x08}[t][t]{6}%
\psfrag{x09}[t][t]{8}%
%
\psfrag{v01}[r][r]{-10}%
\psfrag{v02}[r][r]{-5}%
\psfrag{v03}[r][r]{0}%
\psfrag{v04}[r][r]{5}%
%
\psfrag{z01}[r][r]{0}%
\psfrag{z02}[r][r]{5}%
\psfrag{z03}[r][r]{10}%
\psfrag{z04}[r][r]{15}%
\psfrag{z05}[r][r]{20}%
\psfrag{z06}[r][r]{25}%
%
\resizebox{6cm}{!}{\includegraphics{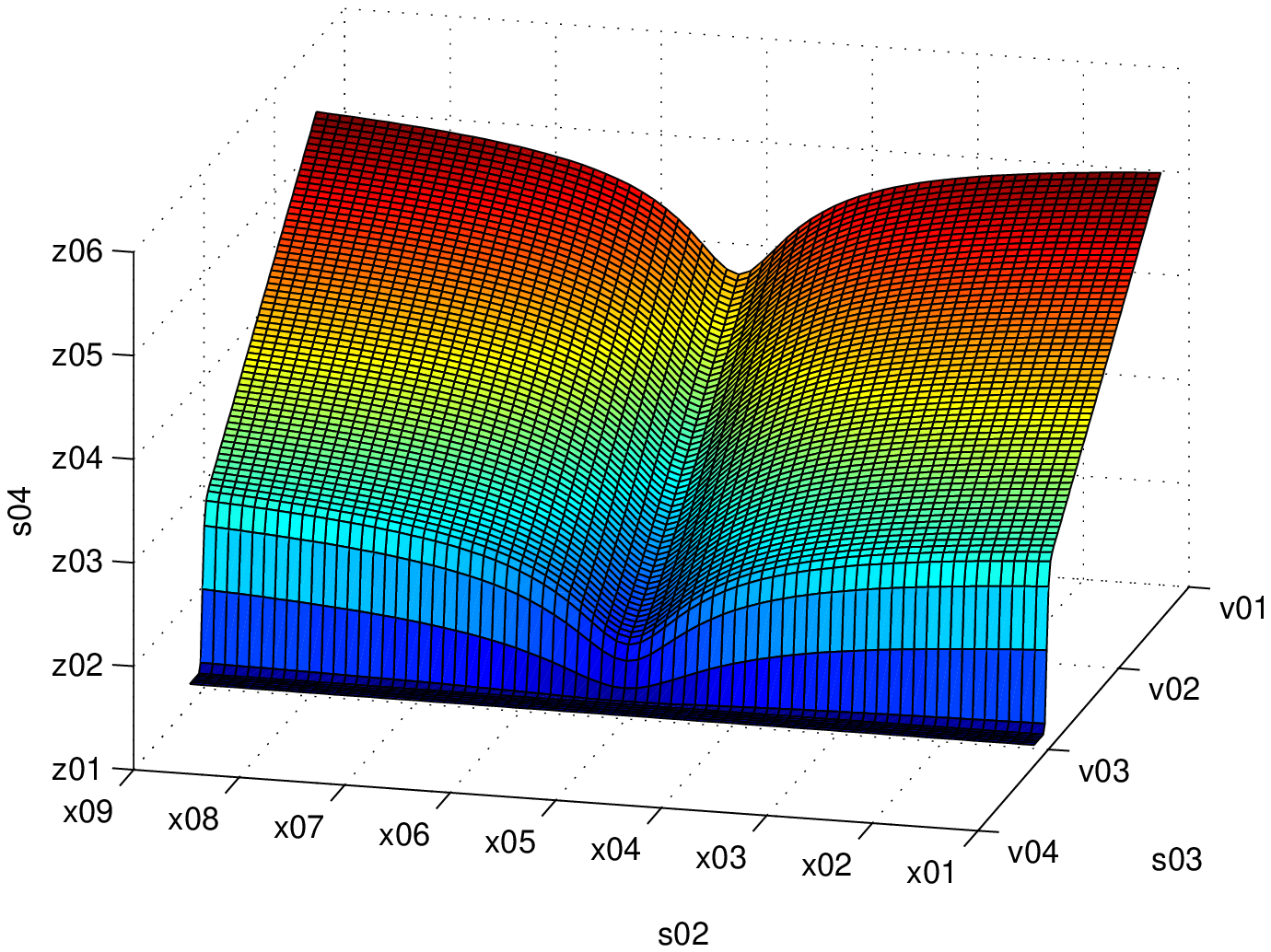}}%
\end{psfrags}%
%

    \end{center}
  \end{minipage}
\caption{Left: $\alpha$ code for rational numbers: Code word lengths increase with $| \theta |$ and $\theta/\delta$. Right: Smooth approximation function for the $\alpha$ coding function.}
\label{alpha}
\end{figure}

In this formula, \texttt{erf} is the error function and $c_0$, $c_1$ and $c_2$ are numerically fitted constants of O(1). The alpha code lengths and their smooth approximations are shown in Figure \ref{alpha}.

\subsection{Spherical coding of a random sequence of signed integers}

We seek a means of coding the part of the data that cannot be explained by the model, i.e. the residuals $\vec{e}$ (a sequence of signed integers). Since only model fitting (compression) can determine that the sequence is not random, we shall build a code assuming that each element is random (incompressible), independent of each other, and uniform (equivalent to an i.i.d sample).

One way we could achieve this is simply to assign a uniform length signed integer code to all possible values ranging from the observed maximum to the observed minimum. This would be equivalent to assuming a uniform error distribution, which would be inefficient as it is not likely in practice. Another possibility would be to assign a universal code to each signed integer, which would be also inefficient, since we would not be exploiting the (reasonable) assumption of uniformity. We chose, therefore, to assume that the likelihood of a sample $\vec{e}$ decreases with its 2-norm (hence the name ``spherical code''), and to use a universal code for coding of each possible $N$-dimensional value in order of increasing 2-norm. This, we will show numerically at least, is equivalent to assuming a Gaussian (normal) distribution of noise.

The coding problem, i.e. a scheme which allows the counting of errors of any size, becomes one of counting the maximal expected number of hypercubes with side lengths $\delta(X)$ that intersect a hypersphere of $N$ dimensions ordered spirally outwards from the origin.

\begin{eqnarray}
H: N \leftrightarrow \mathbb{Z}^N, \ s.t. \left\| H(n) \right\|_2 \leq \left\| H(n+1) \right\|_2 \ \forall n \in N 
\end{eqnarray}

\begin{figure}[htbp]
  \begin{minipage}[t]{.24\textwidth}
    \begin{center}  
%
%
\begin{psfrags}%
\psfragscanon%
%
\psfrag{s05}[t][t]{\color[rgb]{0,0,0}\setlength{\tabcolsep}{0pt}\begin{tabular}{c}\huge log($d$)\end{tabular}}%
\psfrag{s06}[b][b]{\color[rgb]{0,0,0}\setlength{\tabcolsep}{0pt}\begin{tabular}{c}\huge log($r$)\end{tabular}}%
\psfrag{s08}[b][b]{\color[rgb]{0,0,0}\setlength{\tabcolsep}{0pt}\begin{tabular}{c}\huge log(log($h$))\end{tabular}}%
%
\psfrag{x01}[t][t]{0}%
\psfrag{x02}[t][t]{0.1}%
\psfrag{x03}[t][t]{0.2}%
\psfrag{x04}[t][t]{0.3}%
\psfrag{x05}[t][t]{0.4}%
\psfrag{x06}[t][t]{0.5}%
\psfrag{x07}[t][t]{0.6}%
\psfrag{x08}[t][t]{0.7}%
\psfrag{x09}[t][t]{0.8}%
\psfrag{x10}[t][t]{0.9}%
\psfrag{x11}[t][t]{1}%
\psfrag{x12}[t][t]{\Large 0}%
\psfrag{x13}[t][t]{\Large 2}%
\psfrag{x14}[t][t]{\Large 4}%
\psfrag{x15}[t][t]{\Large 6}%
\psfrag{x16}[t][t]{\Large 8}%
\psfrag{x17}[t][t]{\Large 10}%
\psfrag{x18}[t][t]{\Large 12}%
%
\psfrag{v01}[l][l]{}%
\psfrag{v02}[l][l]{}%
\psfrag{v03}[l][l]{}%
\psfrag{v04}[l][l]{}%
\psfrag{v05}[l][l]{}%
\psfrag{v06}[l][l]{}%
\psfrag{v07}[l][l]{}%
\psfrag{v08}[l][l]{}%
\psfrag{v09}[r][r]{\Large 0}%
\psfrag{v10}[r][r]{\Large 1}%
\psfrag{v11}[r][r]{\Large 2}%
\psfrag{v12}[r][r]{\Large 3}%
\psfrag{v13}[r][r]{\Large 4}%
\psfrag{v14}[r][r]{\Large 5}%
\psfrag{v15}[r][r]{\Large 6}%
%
\resizebox{3.2cm}{!}{\includegraphics[clip=true,trim=0 0 15mm 0]{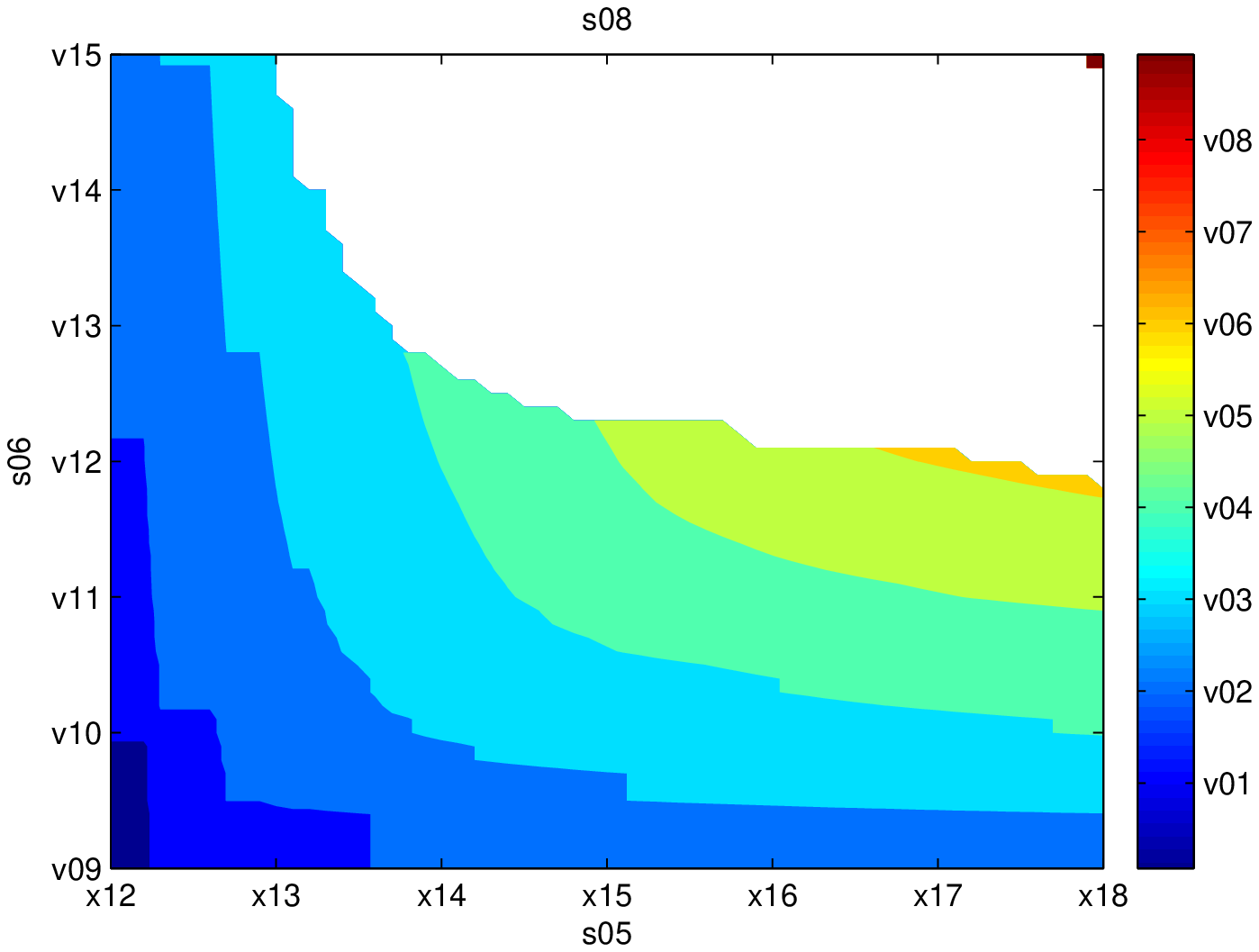}}%
\end{psfrags}%
%

    \end{center}
  \end{minipage}
  \begin{minipage}[t]{.24\textwidth}
    \begin{center}  
%
%
\begin{psfrags}%
\psfragscanon%
%
\psfrag{s05}[t][t]{\color[rgb]{0,0,0}\setlength{\tabcolsep}{0pt}\begin{tabular}{c}\huge log($d$)\end{tabular}}%
\psfrag{s06}[b][b]{\color[rgb]{0,0,0}\setlength{\tabcolsep}{0pt}\begin{tabular}{c} \huge log($r$)\end{tabular}}%
\psfrag{s08}[b][b]{\color[rgb]{0,0,0}\setlength{\tabcolsep}{0pt}\begin{tabular}{c}\huge log(log($V_d(r)$)) \end{tabular}}%
%
\psfrag{x01}[t][t]{0}%
\psfrag{x02}[t][t]{0.1}%
\psfrag{x03}[t][t]{0.2}%
\psfrag{x04}[t][t]{0.3}%
\psfrag{x05}[t][t]{0.4}%
\psfrag{x06}[t][t]{0.5}%
\psfrag{x07}[t][t]{0.6}%
\psfrag{x08}[t][t]{0.7}%
\psfrag{x09}[t][t]{0.8}%
\psfrag{x10}[t][t]{0.9}%
\psfrag{x11}[t][t]{1}%
\psfrag{x12}[t][t]{\Large0}%
\psfrag{x13}[t][t]{\Large2}%
\psfrag{x14}[t][t]{\Large4}%
\psfrag{x15}[t][t]{\Large6}%
\psfrag{x16}[t][t]{\Large8}%
\psfrag{x17}[t][t]{\Large10}%
\psfrag{x18}[t][t]{\Large12}%
%
\psfrag{v01}[l][l]{}%
\psfrag{v02}[l][l]{}%
\psfrag{v03}[l][l]{}%
\psfrag{v04}[l][l]{}%
\psfrag{v05}[l][l]{}%
\psfrag{v06}[l][l]{}%
\psfrag{v07}[l][l]{}%
\psfrag{v08}[l][l]{}%
\psfrag{v09}[r][r]{\Large0}%
\psfrag{v10}[r][r]{\Large1}%
\psfrag{v11}[r][r]{\Large2}%
\psfrag{v12}[r][r]{\Large3}%
\psfrag{v13}[r][r]{\Large4}%
\psfrag{v14}[r][r]{\Large5}%
\psfrag{v15}[r][r]{\Large6}%
%
\resizebox{3.2cm}{!}{\includegraphics[clip=true,trim=0 0 16mm 0]{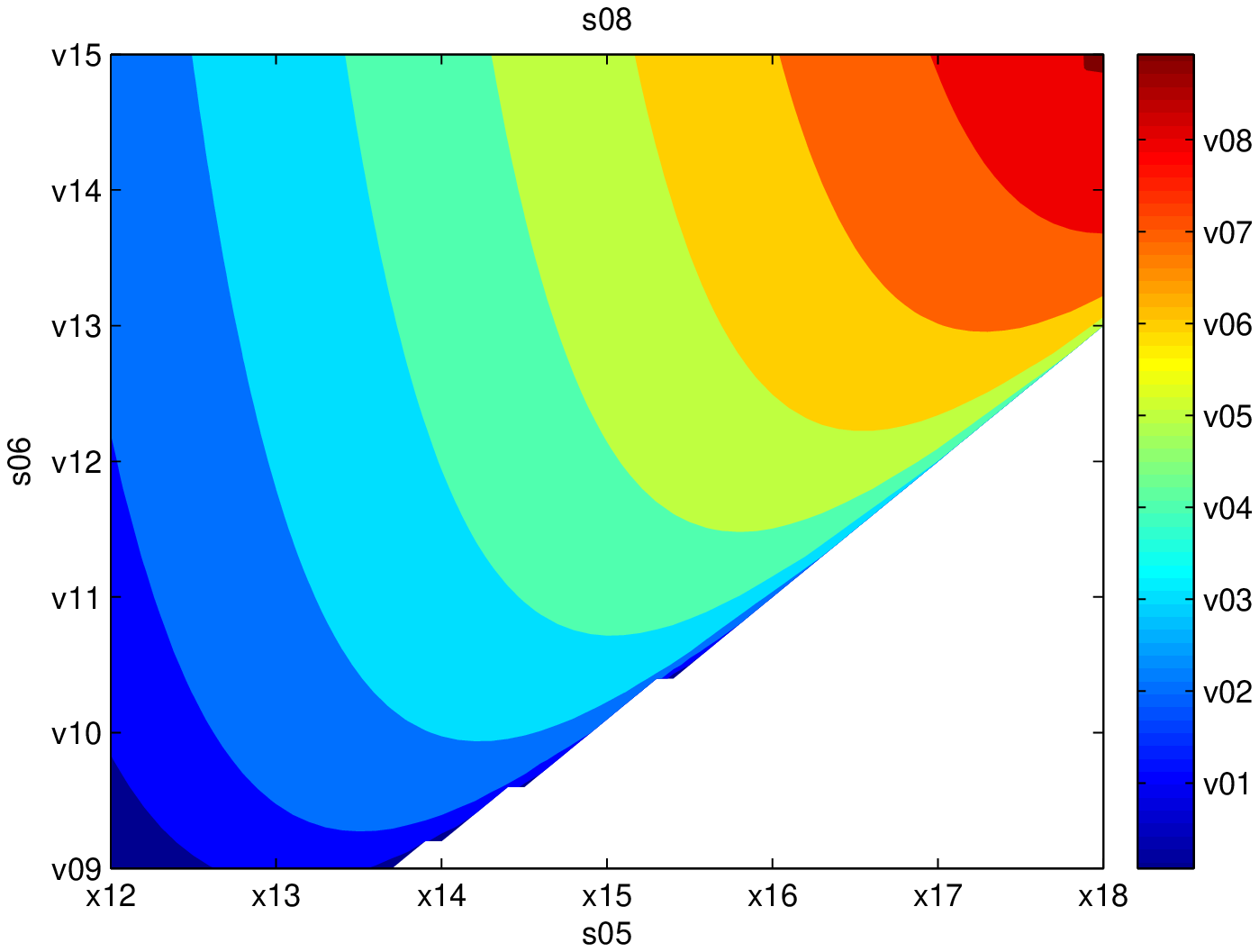}}%
\end{psfrags}%
%

    \end{center}
  \end{minipage}
  \begin{minipage}[t]{.24\textwidth}
    \begin{center}  
%
%
\begin{psfrags}%
\psfragscanon%
%
\psfrag{s05}[t][t]{\color[rgb]{0,0,0}\setlength{\tabcolsep}{0pt}\begin{tabular}{c}\huge log($d$)\end{tabular}}%
\psfrag{s06}[b][b]{\color[rgb]{0,0,0}\setlength{\tabcolsep}{0pt}\begin{tabular}{c}\huge log($r \sqrt{d}$)\end{tabular}}%
\psfrag{s08}[b][b]{\color[rgb]{0,0,0}\setlength{\tabcolsep}{0pt}\begin{tabular}{c}\huge log(log($h_{ap}$))\end{tabular}}%
%
\psfrag{x01}[t][t]{0}%
\psfrag{x02}[t][t]{0.1}%
\psfrag{x03}[t][t]{0.2}%
\psfrag{x04}[t][t]{0.3}%
\psfrag{x05}[t][t]{0.4}%
\psfrag{x06}[t][t]{0.5}%
\psfrag{x07}[t][t]{0.6}%
\psfrag{x08}[t][t]{0.7}%
\psfrag{x09}[t][t]{0.8}%
\psfrag{x10}[t][t]{0.9}%
\psfrag{x11}[t][t]{1}%
\psfrag{x12}[t][t]{\Large 0}%
\psfrag{x13}[t][t]{\Large 2}%
\psfrag{x14}[t][t]{\Large 4}%
\psfrag{x15}[t][t]{\Large 6}%
\psfrag{x16}[t][t]{\Large 8}%
\psfrag{x17}[t][t]{\Large 10}%
\psfrag{x18}[t][t]{\Large 12}%
%
\psfrag{v01}[l][l]{}%
\psfrag{v02}[l][l]{}%
\psfrag{v03}[l][l]{}%
\psfrag{v04}[l][l]{}%
\psfrag{v05}[l][l]{}%
\psfrag{v06}[l][l]{}%
\psfrag{v07}[l][l]{}%
\psfrag{v08}[l][l]{}%
\psfrag{v09}[r][r]{\Large 0}%
\psfrag{v10}[r][r]{\Large 1}%
\psfrag{v11}[r][r]{\Large 2}%
\psfrag{v12}[r][r]{\Large 3}%
\psfrag{v13}[r][r]{\Large 4}%
\psfrag{v14}[r][r]{\Large 5}%
\psfrag{v15}[r][r]{\Large 6}%
%
\resizebox{3.2cm}{!}{\includegraphics[clip=true,trim=0 0 16mm 0]{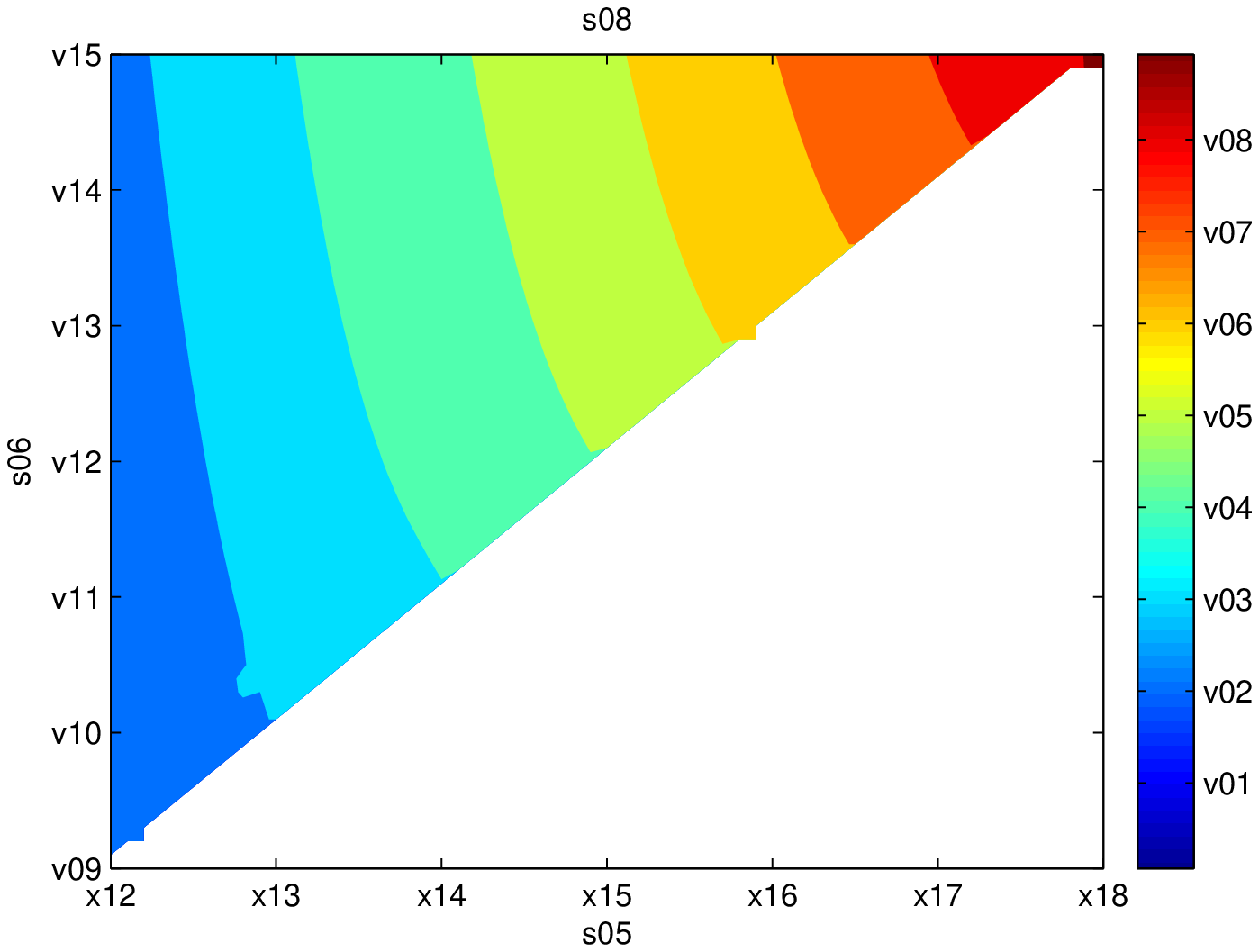}}%
\end{psfrags}%
%

    \end{center}
  \end{minipage}
  \begin{minipage}[t]{.26\textwidth}
    \begin{center}  
%
%
\begin{psfrags}%
\psfragscanon%
%
\psfrag{s05}[t][t]{\color[rgb]{0,0,0}\setlength{\tabcolsep}{0pt}\begin{tabular}{c}\huge log($d$)\end{tabular}}%
\psfrag{s06}[b][b]{\color[rgb]{0,0,0}\setlength{\tabcolsep}{0pt}\begin{tabular}{c}\huge log($r$)\end{tabular}}%
\psfrag{s08}[b][b]{\color[rgb]{0,0,0}\setlength{\tabcolsep}{0pt}\begin{tabular}{c}\huge log(log($\bar{h}$))\end{tabular}}%
%
\psfrag{x01}[t][t]{0}%
\psfrag{x02}[t][t]{0.1}%
\psfrag{x03}[t][t]{0.2}%
\psfrag{x04}[t][t]{0.3}%
\psfrag{x05}[t][t]{0.4}%
\psfrag{x06}[t][t]{0.5}%
\psfrag{x07}[t][t]{0.6}%
\psfrag{x08}[t][t]{0.7}%
\psfrag{x09}[t][t]{0.8}%
\psfrag{x10}[t][t]{0.9}%
\psfrag{x11}[t][t]{1}%
\psfrag{x12}[t][t]{\Large 0}%
\psfrag{x13}[t][t]{\Large 2}%
\psfrag{x14}[t][t]{\Large 4}%
\psfrag{x15}[t][t]{\Large 6}%
\psfrag{x16}[t][t]{\Large 8}%
\psfrag{x17}[t][t]{\Large 10}%
\psfrag{x18}[t][t]{\Large 12}%
%
\psfrag{v01}[l][l]{\Large 1}%
\psfrag{v02}[l][l]{\Large 2}%
\psfrag{v03}[l][l]{\Large 3}%
\psfrag{v04}[l][l]{\Large 4}%
\psfrag{v05}[l][l]{\Large 5}%
\psfrag{v06}[l][l]{\Large 6}%
\psfrag{v07}[l][l]{\Large 7}%
\psfrag{v08}[l][l]{\Large 8}%
\psfrag{v09}[r][r]{\Large 0}%
\psfrag{v10}[r][r]{\Large 1}%
\psfrag{v11}[r][r]{\Large 2}%
\psfrag{v12}[r][r]{\Large 3}%
\psfrag{v13}[r][r]{\Large 4}%
\psfrag{v14}[r][r]{\Large 5}%
\psfrag{v15}[r][r]{\Large 6}%
%
\resizebox{3.6cm}{!}{\includegraphics{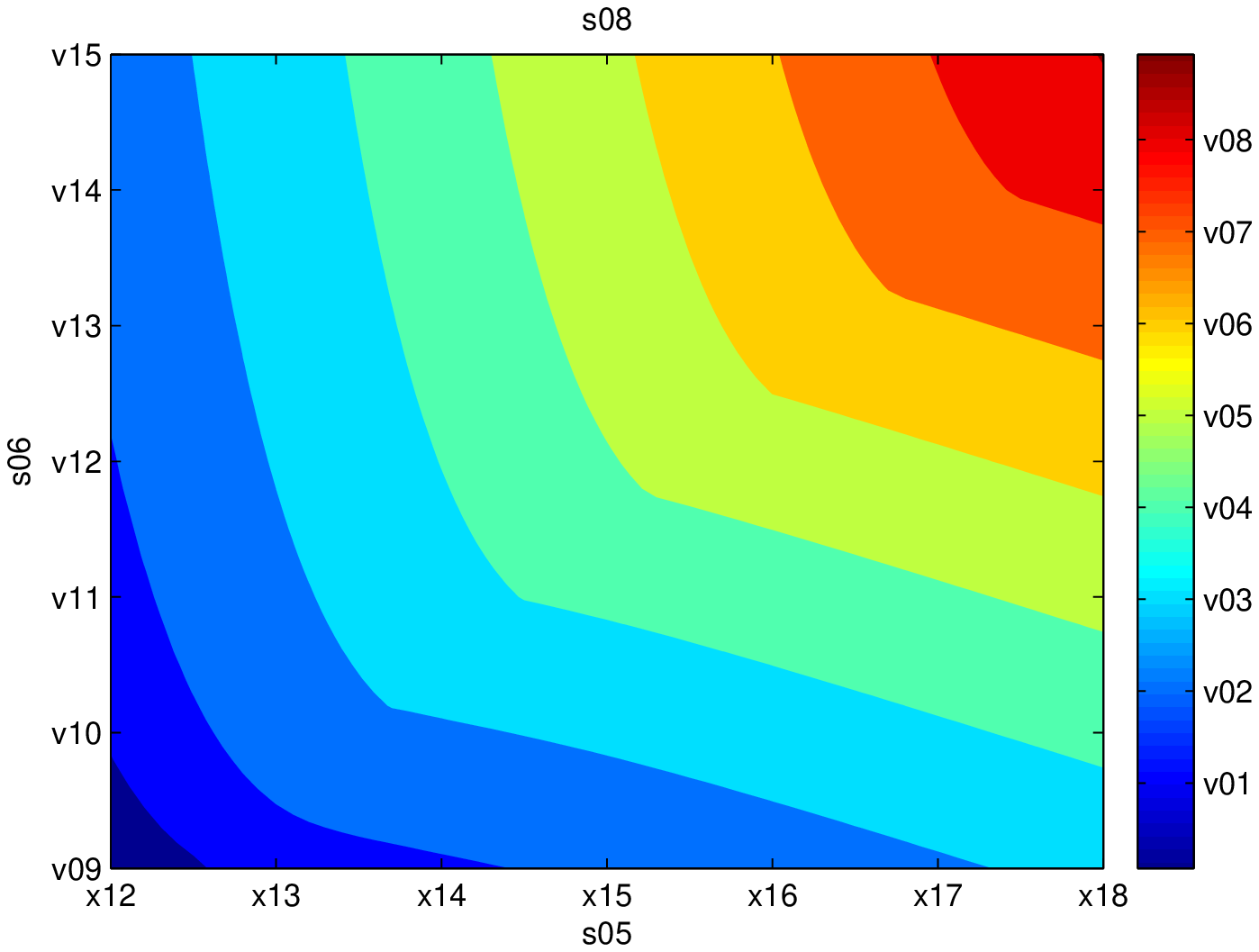}}%
\end{psfrags}%
%

    \end{center}
  \end{minipage}
\caption{Spherical counting scheme and approximations: The first picture shows the surface plot of the exact count $h$. White indicates the region in which computation of $h$ was not feasible. In the second picture we can see the spherical volume. In the lower right region this approximation is not accurate. The third picture shows the Shannon coding length for a spherical data distribution, where we plot $d$ vs. $r/\sqrt{d}$. The reason why this approximation is not valid in the region underneath the first angle bisecting line is that $r/\sqrt{d}$ gets so small that the coding length would be estimated as one bit. The fourth picture shows the approximation function $\bar{h}$.}
\label{happrox}
\end{figure}

So, our spiral counting scheme $H(n)$ guarantees that the 2-norm is monotonically increasing and therefore, since the universal description length of an integer is monotonically increasing with the integer, it follows that under this counting scheme the description length of an integer vector under $H$ is monotonically increasing with its 2-norm. This leads us to state that:

\begin{eqnarray}
\left\|  \frac{\vec{y}}{\delta(X)} \right\|_2 \leq r \Rightarrow K(\vec{y}) = \left| U\left(H^{-1}\left(\frac{\vec{y}}{\delta(X)}\right)\right) \right| \leq h(r,N) \lessapprox \left| U\left(\max\left(V_S^N(r),1\right)\right) \right|, 
\end{eqnarray}

where $V_S^N(r)$ denotes the volume of the hyper-sphere of radius $r$ in $\mathbb{R}^N$. We can establish a connection between $h$ and the Shannon entropy $h_{sh}(x)$. It is the shortest average coding length for a random variable $x$ of dimension $N$. For spherical data with variance $\sigma^2$,

\begin{eqnarray}
 h_{sh}(x) &=& \frac{N}{2} \cdot \log (2 \pi e ) +  N \cdot \log (\sigma)
\end{eqnarray}

To be able to compare $h_{sh}$ and $h$, we have to consider the case of finite data again: We cannot simply assume that $\sigma$ is known, but rather we have to actually encode it and consider its coding length. We do so using the $\alpha$-code. The minimum number of bits that we need to store $\sigma$ can be found by minimizing over the precision $\Delta\sigma$ . We call the resulting \textit{applied} Shannon coding length $h_{ap}$:

\begin{eqnarray}
 h_{ap}(x)  = \frac{N}{2} \cdot \log (2 \pi e ) + \min_{\Delta \sigma} (N \cdot \log (\sigma + \Delta \sigma) + \alpha (\sigma, \Delta \sigma)) \approx h(r = \sigma \sqrt{N}, N) 
\end{eqnarray}

Figure \ref{happrox} shows that the applied Shannon coding length is approximated by the spherical code length function for large $N$. Shannon-Fano or Huffman code lengths/entropies for a given distribution are only valid asymptotically, i.e. for very large $N$. For finite, small $N$, such coding - based on known or sample standard deviation (SD) - may even result in expansion rather than the intended compression. 
\subsection{MDL for linear regression}

The complete (loss-less) code length approximation $\lambda_M$ of the data to be minimized is:

\begin{eqnarray}
\lambda_M(\vec{\theta}, \vec{\delta}, X) = \bar{\alpha}(\vec{\theta}, \vec{\delta}) + \bar{h}(s_M^2(\vec{\theta_{\#}} , X),N),
\end{eqnarray}

where $\vec{\theta_{\#}} = \alpha^{-1}(\alpha(\vec{\theta},\vec{\delta}))$. Bars on $\alpha$ and $h$ denote smooth approximations. Note that this equation requires the function $s_M(\cdot)$, which relates the size of the residual 2-norm to the parameters and their storage precision, for some model $M$ which specifies how the parameters encode $\vec{y}$, usually represented by a Hessian function. This principle can be applied to any type of regression fit: we shall do so for linear regression. 

Finding a close, smooth bound $\bar{h}$ for any $r$ and $N$ is a difficult mathematical problem, which also can be formulated as finding the number of representations of an integer as a sum of squares. A general (approximative) function has not been found yet \cite{grosswald85}. For combinations of large enough $r$ ($>5$) and low enough $N$ ($<10^6$) (see previous section) the following approximation based on the unit sphere volume is reasonable:

\begin{eqnarray}
\bar{h}(s^2(\vec{e})) = \log \left( \frac{\pi^\frac{N}{2}}{\Gamma \left( \frac{N}{2} + 1 \right) } \right) + \frac{N}{2} \log \left( \frac{s^2(\vec{\theta_{\#}}, X)}{\delta(X)^2} \right)
\end{eqnarray}

The $\alpha$ code has a rather nice property in that the numerical value of the decoding $\alpha^{-1}(\alpha(\theta_i,\delta_i)) = \theta_i + \delta(\theta_i)$ is uniformly distributed on the interval $[\theta_i - \delta_i, \theta_i + \delta_i]$ with $p_i(\delta)=(2\delta_i)^{-1}$, over all possible pairs of $\theta$ and $\delta$: as Monte-Carlo simulations have confirmed, the code behaves without bias of $((\theta_{\#})_i - \theta_i)/\delta_i$. This allows us to take the expected value of the increase in the norm of the residual for random, independent perturbations $\delta(\theta_i)$ of each $\theta_i$ with the uniform probability $p_i(\delta)$. Calculating the growth of residual norm (squared) with respect to parameter accuracy $\vec{\delta}$ is straightforward for linear regression:

\begin{eqnarray}
s^2(\alpha^{-1}(\alpha(\vec{\theta},\vec{\delta})),X) &=& \left( \vec{y} - X \cdot \alpha^{-1}(\alpha(\vec{\theta},\vec{\delta})) \right)^T \cdot \left( \vec{y} - X \cdot \alpha^{-1}(\alpha(\vec{\theta},\vec{\delta})) \right) \nonumber \\
&=& \left( \vec{y} - X \cdot ( \vec{\theta} + \vec{\delta(\theta)} ) \right)^T \cdot \left( \vec{y} - X \cdot ( \vec{\theta} + \vec{\delta(\theta)} ) \right)
\end{eqnarray}

We choose $\vec{\theta} = \left( X^T X \right)^{-1} X^T y$ to minimize the 2-norm of the residual $ \vec{e}^T\vec{e}$, (or any exact solution in the underdetermined case) so the gradient is with respect to $\vec{\theta}$ zero, meaning that any perturbation from $\vec{\theta}$ results in strictly quadratic growth: $ \vec{e}^T\vec{e} + \vec{\delta(\theta)}^T X^T X \vec{\delta(\theta)} = \vec{e}^T\vec{e} + \vec{\delta(\theta)}^T \Sigma_X \vec{\delta(\theta)} $, where $\Sigma_X$ is the covariance of $X$. Taking the expected value over possible perturbations (i.e. the outputs of the coder-decoder): 

\begin{eqnarray}
E \left[ \delta(\theta_i)^T (\Sigma_X)_{i,i} \delta(\theta_i) \right] &=& \int_{-\delta_i}^{\delta_i} p(\delta(\theta_i)) (\Sigma_X)_{i,i} \delta(\theta_i)^2 d\delta(\theta_i) \nonumber\\
&=& (2 \delta_i)^{-1} \frac{2}{3} (\Sigma_X)_{i,i} \delta_i^3 = \frac{1}{3} (\Sigma_X)_{i,i} \delta_i^2 \nonumber\\
E \left[ \delta(\theta_i)^T (\Sigma_X)_{i,j} \delta(\theta_j) \right]_{i \neq j} &=& c \int_{-\delta_j}^{\delta_j} \int_{-\delta_i}^{\delta_i} (\Sigma_X)_{i,j} \delta(\theta_i) \delta(\theta_j) d\delta(\theta_i) d\delta(\theta_j) = 0 \nonumber \\
E \left[ \delta(\vec{\theta})^T \Sigma_X \delta(\vec{\theta}) \right] &=& E \left[ \sum_{i,j} \delta(\theta_i)^T (\Sigma_X)_{i,j} \delta(\theta_j) \right] = \frac{1}{3} \sum_i (\Sigma_X)_{i,i} \delta_i^2
\end{eqnarray}

Finally, a differentiable CLR objective function for linear regression can be written.

\begin{eqnarray}
\min_{\vec{\theta}} \left( E_{\delta(\theta)} \left[ \lambda_M(\vec{\theta}, \vec{\delta}, X) \right] \right) &=& \min_{\vec{\theta}} \left( \bar{\alpha}(\vec{\theta}, \vec{\delta}) + E_{\delta(\theta)} \left[ \bar{h}(s_M^2 + \delta(X)^T \Sigma_{X,M} \delta(X),N) \right] \right) \nonumber\\
&\approx& \min_{\vec{\theta}} \left( \sum_i \bar{\alpha}(\theta_i, \delta_i) +  \bar{h}(s_M^2 + \frac{1}{3} \sum_i (\Sigma_X)_{i,i} \delta_i^2,N) \right)
\end{eqnarray}

The approximation $E[h(a+x)]=E[h(a)+h^\prime(a)x+...]=h(a)+h^\prime(a)E[x]+... \cong h(a+E[x])$ is valid if $h(a)$ is locally linear - being $O(\log(a))$ that assumption is reasonable. The resulting objective function is smooth in optimization parameters ($2K$ such parameters: $\vec{\theta}$ and $\vec{\delta}$) but is non-convex. A useful unbiased heuristic we have found was to use simplex optimization (code was written in Matlab, Mathworks, Inc.) with starting value for $\vec{\theta}$ at the usual 2-norm solution and $\delta_i = |\theta_i|/2$. After each downhill optimization, all parameters for which $\delta_i > |\theta_i|$ were discarded and the entire procedure repeated (with reduced X) until no further parameters were thus "`culled"'.

\section{Results}

The CLR method proposed was compared to 2 different LASSO variants on both simulated and real data. Simulated data allows us to gauge the relative ability of CLR to accurately recover the sparsity structure in the case in which it is known. We produced the same synthetic data sets as did Tibshirani in his seminal LASSO paper \cite{tibshirani96}.  For each of three examples, we simulated 50 data sets consisting of 20 observations from the model $y = \vec{\theta}^T \vec{x} + \sigma \vec{\epsilon} $, where $\epsilon$ is standard normal. The correlation between $x_i$ and $x_j$ was $\rho^{|i-j|}$ with $\rho = 0.5$. The dataset paramters for SIM1 were $\vec{\beta{}} = (3, 1.5, 0, 0, 2, 0, 0, 0)^T$ and $\sigma=3$, for SIM2 $\beta_j = 0.85, \forall j$ and $\sigma=3$ and for SIM3 $\beta=(5,0,0,0,0,0,0,0)$ and $\sigma=2$. In \cite{tibshirani96} LASSO performed favorably compared to other methods such as ridge regression and therefore only LASSO and normal 2-norm regression was performed for comparison. Performance results are given in Table 1. for each of the simulated datasets (SIM1, SIM2, SIM3) and methods evaluated (CLR - parsimonious linear regression, LASSOLCR - LASSO largest consistent region, LASSOCV - LASSO with cross-validation, similar to Tibshirani's cross-validated LASSO in \cite{tibshirani96} which was found to be highly performant). Both variants used bootstrapping for LASSO statistic estimation. We used the LASSO implementation available in the Matlab repository (www.mathworks.com/matlabcentral/fileexchange) written by M. Dunham and described in \cite{bach08}. The given values are mean running time (as a fraction of L2 running time) over all simulations, and for each dataset: the number of non-zero parameters computed, the minimum squared error (MSE) of the computed parameters vs. simulated parameters and the square root of the MSE of residuals as a fraction of the size of the noise used for the simulation $\sigma$. The bias value is not included in these calculations. For L2 the full parameters set is always returned and therefore 8 vs. 2 means that 2 of the actual parameters for that dataset were non-zero. 

\begin{table}[ht]
\caption{Results of sparse linear regressions on simulated datasets.}
\begin{center}
\begin{tabular}{|l|l|c|c|c|c|}
\hline
\textbf{Method} & &\textbf{CLR}&\textbf{LASSOLCR}&\textbf{LASSOCV}&\textbf{L2}\\\hline
\textbf{Time} & \textbf{}&112&204&202&1\\\hline
\multirow{3}{*}{\textbf{ nr. of parameters found}} & \textbf{SIM1}&2.10$\pm$1.4&0.32$\pm$0.5&3.40$\pm$2.2&8 vs. 3\\
& \textbf{SIM2}&1.26$\pm$1.1&0.04$\pm$0.2&4.08$\pm$1.7&8 vs. 8\\
& \textbf{SIM3}&1.72$\pm$1.3&0.90$\pm$0.3&2.68$\pm$2.0&8 vs. 1\\\hline
\multirow{3}{*}{\textbf{MSE}} & \textbf{SIM1}&1.27$\pm$0.8&1.72$\pm$0.4&1.01$\pm$0.6&1.10$\pm$0.8\\
& \textbf{SIM2}&1.12$\pm$0.4&0.74$\pm$0.1&1.10$\pm$0.6&1.04$\pm$0.8\\
& \textbf{SIM3}&0.19$\pm$0.4&0.33$\pm$0.9&0.25$\pm$0.4&0.72$\pm$0.8\\\hline
\multirow{3}{*}{\textbf{$s_d(\vec{e})/\sigma$}} & \textbf{SIM1}&1.05$\pm$0.3&1.68$\pm$0.4&0.91$\pm$0.2&0.73$\pm$0.1\\
& \textbf{SIM2}&1.28$\pm$0.4&1.77$\pm$0.3&0.91$\pm$0.2&0.75$\pm$0.1\\
& \textbf{SIM3}&0.94$\pm$0.2&1.11$\pm$0.4&0.87$\pm$0.2&0.75$\pm$0.2\\\hline
\end{tabular}
\end{center}
\end{table}

\begin{figure}[ht]
  \begin{center}  
%
%
\begin{psfrags}%
\psfragscanon%
%
\psfrag{s05}[t][t]{\color[rgb]{0,0,0}\setlength{\tabcolsep}{0pt}\begin{tabular}{c}$SD(\vec{e})_{test} / SD(\vec{y})_{test}$ \end{tabular}}%
\psfrag{s06}[b][b]{\color[rgb]{0,0,0}\setlength{\tabcolsep}{0pt}\begin{tabular}{c}$SD(\vec{e})_{training} / SD(\vec{y})_{training}$ \end{tabular}}%
\psfrag{s10}[][]{\color[rgb]{0,0,0}\setlength{\tabcolsep}{0pt}\begin{tabular}{c} \end{tabular}}%
\psfrag{s11}[][]{\color[rgb]{0,0,0}\setlength{\tabcolsep}{0pt}\begin{tabular}{c} \end{tabular}}%
\psfrag{s12}[l][l]{\color[rgb]{0,0,0}\setlength{\tabcolsep}{0pt}\begin{tabular}{l}1\end{tabular}}%
\psfrag{s13}[l][l]{\color[rgb]{0,0,0}\setlength{\tabcolsep}{0pt}\begin{tabular}{l}3\end{tabular}}%
\psfrag{s14}[l][l]{\color[rgb]{0,0,0}\setlength{\tabcolsep}{0pt}\begin{tabular}{l}4\end{tabular}}%
\psfrag{s15}[l][l]{\color[rgb]{0,0,0}\setlength{\tabcolsep}{0pt}\begin{tabular}{l}5\end{tabular}}%
\psfrag{s16}[l][l]{\color[rgb]{0,0,0}\setlength{\tabcolsep}{0pt}\begin{tabular}{l}6\end{tabular}}%
\psfrag{s17}[l][l]{\color[rgb]{0,0,0}\setlength{\tabcolsep}{0pt}\begin{tabular}{l}7\end{tabular}}%
\psfrag{s18}[l][l]{\color[rgb]{0,0,0}\setlength{\tabcolsep}{0pt}\begin{tabular}{l}4\end{tabular}}%
\psfrag{s19}[l][l]{\color[rgb]{0,0,0}\setlength{\tabcolsep}{0pt}\begin{tabular}{l}3\end{tabular}}%
\psfrag{s20}[l][l]{\color[rgb]{0,0,0}\setlength{\tabcolsep}{0pt}\begin{tabular}{l}4\end{tabular}}%
\psfrag{s21}[l][l]{\color[rgb]{0,0,0}\setlength{\tabcolsep}{0pt}\begin{tabular}{l}5\end{tabular}}%
\psfrag{s22}[l][l]{\color[rgb]{0,0,0}\setlength{\tabcolsep}{0pt}\begin{tabular}{l}6\end{tabular}}%
\psfrag{s23}[l][l]{\color[rgb]{0,0,0}\setlength{\tabcolsep}{0pt}\begin{tabular}{l}7\end{tabular}}%
\psfrag{s24}[l][l]{\color[rgb]{0,0,0}\setlength{\tabcolsep}{0pt}\begin{tabular}{l}2\end{tabular}}%
\psfrag{s25}[l][l]{\color[rgb]{0,0,0}\setlength{\tabcolsep}{0pt}\begin{tabular}{l}2\end{tabular}}%
\psfrag{s26}[l][l]{\color[rgb]{0,0,0}\setlength{\tabcolsep}{0pt}\begin{tabular}{l}3\end{tabular}}%
\psfrag{s27}[l][l]{\color[rgb]{0,0,0}\setlength{\tabcolsep}{0pt}\begin{tabular}{l}5\end{tabular}}%
\psfrag{s28}[l][l]{\color[rgb]{0,0,0}\setlength{\tabcolsep}{0pt}\begin{tabular}{l}6\end{tabular}}%
\psfrag{s29}[l][l]{\color[rgb]{0,0,0}\setlength{\tabcolsep}{0pt}\begin{tabular}{l}7\end{tabular}}%
\psfrag{s30}[l][l]{\color[rgb]{0,0,0}\setlength{\tabcolsep}{0pt}\begin{tabular}{l}2\end{tabular}}%
\psfrag{s31}[l][l]{\color[rgb]{0,0,0}\small LASSOCV}%
\psfrag{s32}[l][l]{\color[rgb]{0,0,0}\small PLR}%
\psfrag{s33}[l][l]{\color[rgb]{0,0,0}\small LASSOLCR}%
\psfrag{s34}[l][l]{\color[rgb]{0,0,0}\small LASSOCV}%
%
\psfrag{x01}[t][t]{0}%
\psfrag{x02}[t][t]{0.1}%
\psfrag{x03}[t][t]{0.2}%
\psfrag{x04}[t][t]{0.3}%
\psfrag{x05}[t][t]{0.4}%
\psfrag{x06}[t][t]{0.5}%
\psfrag{x07}[t][t]{0.6}%
\psfrag{x08}[t][t]{0.7}%
\psfrag{x09}[t][t]{0.8}%
\psfrag{x10}[t][t]{0.9}%
\psfrag{x11}[t][t]{1}%
\psfrag{x12}[t][t]{0}%
\psfrag{x13}[t][t]{0.5}%
\psfrag{x14}[t][t]{1}%
\psfrag{x15}[t][t]{1.5}%
\psfrag{x16}[t][t]{2}%
\psfrag{x17}[t][t]{2.5}%
%
\psfrag{v01}[r][r]{0}%
\psfrag{v02}[r][r]{0.1}%
\psfrag{v03}[r][r]{0.2}%
\psfrag{v04}[r][r]{0.3}%
\psfrag{v05}[r][r]{0.4}%
\psfrag{v06}[r][r]{0.5}%
\psfrag{v07}[r][r]{0.6}%
\psfrag{v08}[r][r]{0.7}%
\psfrag{v09}[r][r]{0.8}%
\psfrag{v10}[r][r]{0.9}%
\psfrag{v11}[r][r]{1}%
\psfrag{v12}[r][r]{-0.4}%
\psfrag{v13}[r][r]{-0.2}%
\psfrag{v14}[r][r]{0}%
\psfrag{v15}[r][r]{0.2}%
\psfrag{v16}[r][r]{0.4}%
\psfrag{v17}[r][r]{0.6}%
\psfrag{v18}[r][r]{0.8}%
\psfrag{v19}[r][r]{1}%
\psfrag{v20}[r][r]{1.2}%
\psfrag{v21}[r][r]{1.4}%
%
\resizebox{8cm}{!}{\includegraphics{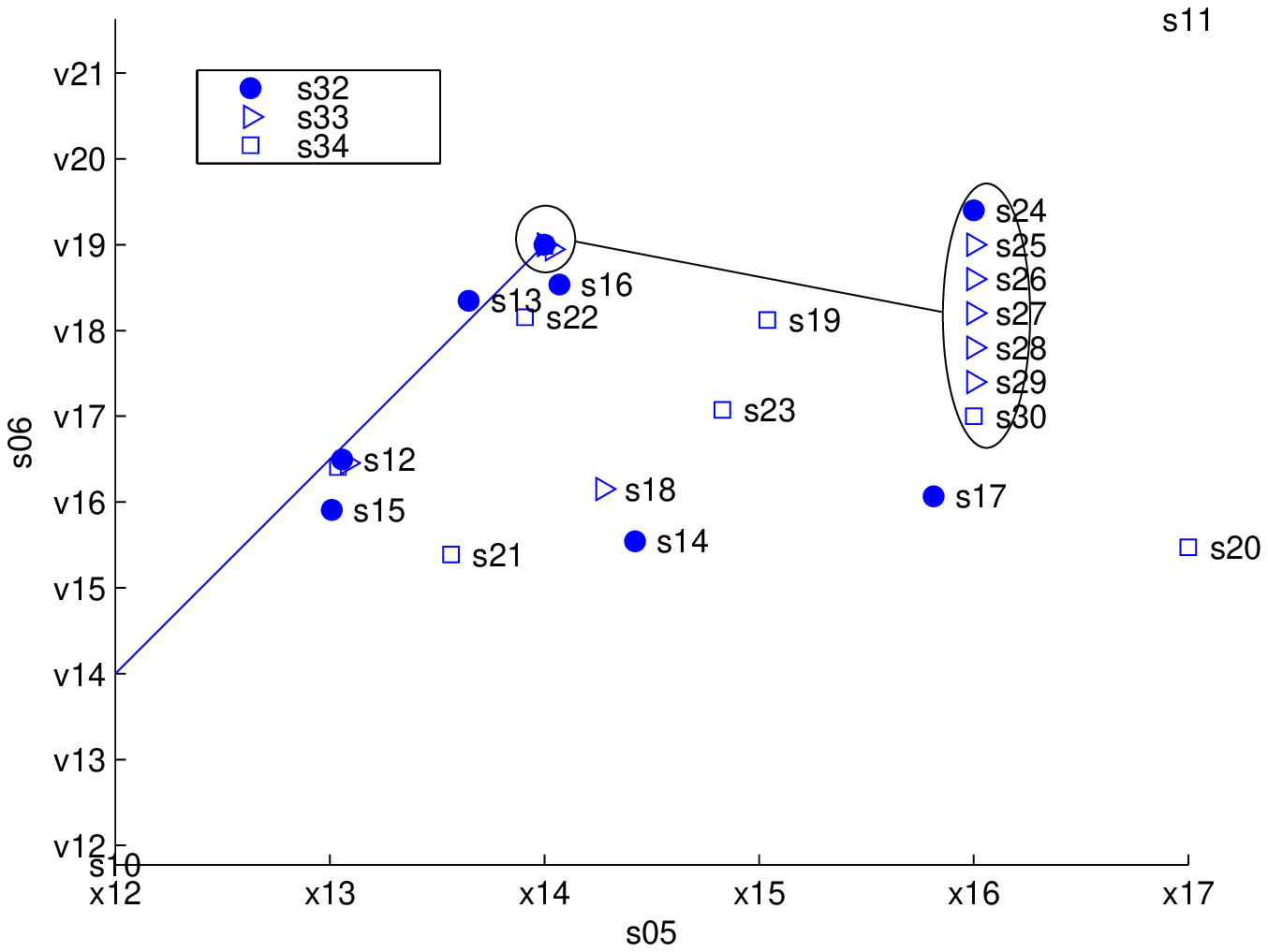}}%
\end{psfrags}%
%

  \end{center}
\caption{Generalization performance on real data. The datasets are numbered as follows: 1. Concrete (N=1030, J=8), 2. Forest Fires (N=517, J=12), 3. Detroit (N=60, J=15), 4. Housing (N=506, J=13), 5. Automobile (N=159, J=15), 6. Triazines (N=186, J=31), 7. Pyrimidines (N=74, J=26).}
\label{results_figure}
\end{figure}

In order to test generalization performance, we used the following datasets from the UCI repository \cite{asuncion07}: Automobile, Concrete Compressive Strength \cite{yeh98}, Forest Fires \cite{cortez07}, Housing, Qualitative Structure Activity Relationships (Pyrimidines and Triazines). We also used one more dataset from the Statlib Archives (http://lib.stat.cmu.edu/): Detroit \cite{gunst80}. For generalization estimation, we used a training test split of 2/3 to 1/3. A feature product of $\text{X}_{k+1+i}=D_{i}^{2}$ was used along with a bias term for all methods. Figure \ref{results_figure} shows training vs. test set errors on the 7 datasets evaluated. The axes are standard deviations of residuals as a fraction of the standard deviation of the target, for both test (x-axis) and training subsets (y-axis). Note that many analyses (especially LASSOLCR) resulted in no features (except bias) being selected, i.e. the ratio of s.d.'s equaled 1. The callout in the top right corner is a list of such instances. The actual test s.d. ratio value for LASSOCV, dataset 4, was 59.0 (off scale). The solution sparsity, averaged over all datasets, was $16.7\%$ for CLR, $11.7\%$ for LASSOLCR and $37.4\%$ for LASSOCV.

\section{Conclusion}

In this paper, we have outlined both the theoretical justification for a detailed MDL application to linear regression which concurrently provides both regularization, feature selection and model selection. At first glance, the complex derivation may appear to be a bit of an overkill. However, simpler MDL applications to the feature selection problem, which have been previously proposed (MDL as a model-selection is long in use for time-series linear auto-regressive modeling), do not perform as well to our experience. Why so? First of all because code length estimates based on Shannon entropy do not work well for small numbers of examples (hence the meticulous derivation of the spherical code). The dataset results we present herein are precisely for small $N$ problems, where there is not much to compress. In the case of auto-regressive time series modeling, it was previously shown that a simpler version of our method exhibits increasing sparsity structure recovery accuracy with increasing $N$. The same non-standard MDL/MML situation applies for large values of quantization width $\delta$ with respect to standard deviation of the target. In fact, $\delta$ can be (and was) inferred directly from the histogram of the target. Histograms of the modelling errors (residuals) on the ``real'' datasets showed that only one of these passed the Jarque-Berra test for normality (Detroit) yet CLR worked well for 2 other datasets. CLR did not perform well on 2 datasets, but neither did either LASSO variant. Note also that LASSO-CV highly overfit on one of the datasets, highlighting the overfit risk of cross-validation. Overall CLR was quite conservative, and it outperformed LASSO on the sparse simulated datasets (in avoiding overfit and recovering the structure). If there is any generally discernible trend in CLR, it is toward underfit, which is not suprising since we are restricting ourselves to a smaller class of models than that from which the 'generating' model may have belonged to.

The CLR method is so conservative, in fact, that it also ``works'' in the underdetermined case, not only allowing us to use product features such as the square but also all combinations of feature product pairs, allowing for a full multi-dimensional polynomial fit. It may be surprising to note that CLR actually worked faster than the ``convex'' LASSO although it is not convex. This is because of the reliance on cross-validation and bootstrapping that hyperparameter choice traditionally requires, both in LASSO and other classification/regression paradigms. As the number of features increases, the relative running time of gradient descent algorithms will increase relative to the LASSO, and further work will be necessary to provide for convex approximation to the MDL and customized, more efficient optimization algorithms. As it is known that feature selection is, per se, a non-convex, NP-complete problem \cite{davies94} it is not likely that the computational time burden will decrease without additional compromises. We'd like to assert, credibly, that there are no current ML approaches which are fully convex, either because of the hyperparameter choice or the adjustment of prior variances. However, our work points to the possibility of fully automatic learning, with no practitioner imposed choices necessary. We have already implemented heuristic algorithms which iteratively generate, and then cull/prune features to minimize MDL, which is an unbounded-time formulation. Since learning in general, at least under the algorithmic information theory perspective, is incomputable, it should not be surprising that heuristic approaches with practically imposed time bounds will be necessary. While the common Gaussian i.i.d assumption was used herein, the MDL formulation allows for learning under other, or even unkown, noise sources. The underlying philosophy of our ML approach is to spend computational resources on model generation and selection rather than validation.

\nocite{*}
\bibliographystyle{plain}
\bibliography{mdl_paper_nips2009}

\section{Acknowledgements}

We thank Prof. Ulf Leser (Humboldt University Berlin) for his helpful comments.

This manuscript was submitted to the Neural Information Processing Conference (NIPS) 2009.

\end{document}